\documentclass[12pt]{article}

\usepackage{amsmath,amsthm, amsfonts, amssymb, amsxtra, amsopn}
\usepackage{pgfplots}
\usetikzlibrary{calc}
\pgfplotsset{compat=1.13}
\usepackage{pgfplotstable}
\usepackage{graphicx,grffile}
\usepackage{multirow}
\usepackage{float}
\usepackage{booktabs}
\usepackage{listings}
\usepackage{cmap}
\usepackage{colortbl}
\usepackage{adjustbox}
\usepackage{epsfig}
\usepackage{bold-extra}

\usepackage[tableposition=top,font=small,skip=5pt]{caption}
\usepackage{subcaption}
\usepackage{makecell} 

\def\cO{{\cal O}}

\PassOptionsToPackage{hyphens}{url}

\usepackage{hyperref}
\hypersetup{colorlinks=true,linkcolor=black,citecolor=black,urlcolor=blue,filecolor=black}
\hypersetup{pdfpagemode=UseNone,pdfstartview=}

\usepackage{enumitem}
\setlist[itemize]{noitemsep, topsep=0pt}

\advance\oddsidemargin by -0.35in
\advance\textwidth by 0.7in

\advance\topmargin by -0.4in
\advance\textheight by 0.8in

\long\def\symbolfootnotetext[#1]#2{\begingroup%
\def\thefootnote{\fnsymbol{footnote}}\footnotetext[#1]{#2}\endgroup}

\DeclareMathOperator{\thth}{th}

\let\vvv=\v
\def\v{{\tt v}}

\def\cO{{\cal O}}

\def\zz{\phantom{0}}

\title{Malware Classification using a Hybrid Hidden Markov Model-Convolutional Neural Network}

\author{Ritik Mehta\footnotemark[1]\ \ \ 
Olha  Jure\vvv{c}kov\'{a}\footnotemark[2]\ \ \ \
Mark Stamp\footnotemark[1]\,\,\footnotemark[3]}

\begin{document}

\symbolfootnotetext[1]{Department of Computer Science, San Jose State University}
\symbolfootnotetext[2]{Faculty of Information Technology, Czech Technical University in Prague}
\symbolfootnotetext[3]{mark.stamp$@$sjsu.edu}

\maketitle

\abstract
The proliferation of malware variants poses a significant challenges to traditional malware detection approaches, 
such as signature-based methods, necessitating the development of advanced machine learning techniques. 
In this research, we present a novel approach based on a hybrid architecture combining features extracted 
using a Hidden Markov Model (HMM), with a Convolutional Neural Network (CNN) then used for 
malware classification. Inspired by the strong results in previous 
work using an HMM-Random Forest model, we propose integrating HMMs, which serve to capture sequential patterns in 
opcode sequences, with CNNs, which are adept at extracting hierarchical features. We demonstrate the effectiveness of 
our approach on the popular Malicia dataset, and we obtain superior performance, as compared to other machine learning 
methods---our results surpass the aforementioned HMM-Random Forest model. Our findings underscore the potential of 
hybrid HMM-CNN architectures in bolstering malware classification capabilities, offering several promising avenues for further 
research in the field of cybersecurity.

\section{Introduction}

Malicious software, commonly known as malware, poses a significant threat to computer systems by causing damage or disruption. 
Despite advancements in cybersecurity, malware continues to present a formidable challenge in the digital landscape. For example,
ransomware attacks increased by~84\% in~2023 as compared to~2022, according to a study conducted by the
NCC Group~\cite{ncc}. This escalating trend underscores the urgent need for improved methods of detecting and 
categorizing malware.

Traditional signature-based techniques, as employed by anti-virus (AV) applications~\cite{wolpin}, entail creating signatures 
consisting of patterns extracted from malicious software files. However, these techniques are ineffective against previously-unknown 
malware samples, and numerous code obfuscation techniques~\cite{obfuscationTechniques} have been developed that can defeat 
signature scans. In contrast, heuristic analysis~\cite{heuristicanalysis} requires careful calibration to balance threat identification with 
excessive false positive rates on benign code.

Recognizing the limitations of these conventional methods, researchers have turned to machine learning paradigms for solutions. 
In this regard, static and dynamic features, or a combination of the two~\cite{Anusha}, are used to train models for malware detection 
and classification. Static features are those which can be obtained without executing or emulating the code, while dynamic features 
require code execution or emulation. In general, models that rely on static features are more efficient as such features are easy to 
extract and have low computation complexity, while models that use dynamic features are more resistant to common obfuscation 
techniques. In our research, we only consider static features.

In this paper, we propose a novel hybrid machine learning technique, termed HMM-CNN, which combines the sequential insight of 
Hidden Markov Model (HMM)~\cite{hmm} with the spatial awareness of Convolutional Neural Networks (CNN). Specifically,
we first train HMMs on opcode sequences, then we determine the hidden state sequences from the trained HMMs. This use
of HMMs can be viewed as a feature engineering step, and it is often employed in the field of Natural Language Processing (NLP),
but we are not aware of such an approach having been previously used in the malware domain. Finally, we 
classify malware samples into their respective families based on these HMM-generated hidden state sequences using a CNN. 
This study is an extension of our prior research efforts that culminated in the development of the 
HMM-Random Forest model in~\cite{hmm-rf}.

The remainder of this paper is organized as follows. In Section~\ref{sect:back} we 
present relevant background information and a brief introduction to the learning techniques considered in our research.
Section~\ref{sect:review} presents a selective survey of some relevant previous work. Section~\ref{sect:meth} covers our 
experimental design, with the emphasis on our proposed HMM-CNN model, and we provide a brief description of the dataset used. Section~\ref{sect:exp} gives our experimental results. We conclude the paper with Section~\ref{sect:conc}, 
which includes some ideas for future work.

\section{Background}\label{sect:back}

In this section, we first introduce the learning techniques that appear in subsequent sections of this paper.
Specifically, we discuss Hidden Markov Models and Convolutional Neural Networks.

\subsection{Hidden Markov Model}

\begin{figure}[!htb]
\centering
\adjustbox{scale=1.0}{
    \begin{tikzpicture}[scale=1.0]
    
    \draw[thick,color=blue] (0,0) rectangle (1,1);
    \draw[thick,color=blue] (2.5,0) rectangle (3.5,1);
    \draw[thick,color=blue] (5,0) rectangle (6,1);
    \draw[thick,color=blue] (10,0) rectangle (11,1);

    \draw[thick,color=green] (0.5,4.5) circle (0.575);
    \draw[thick,color=green] (3,4.5) circle (0.575);
    \draw[thick,color=green] (5.5,4.5) circle (0.575);
    \draw[thick,color=green] (10.5,4.5) circle (0.575);
    
    \node at (0.5,0.5){$\cO_0$};
    \node at (3,0.5){$\cO_1$};
    \node at (5.5,0.5){$\cO_2$};
    \node at (8,0.5){$\cdots$};
    \node at (10.5,0.5){$\cO_{T-1}$};

    \node at (0.5,4.5){$X_0$};
    \node at (3,4.5){$X_1$};
    \node at (5.5,4.5){$X_2$};
    \node at (8,4.5){$\cdots$};
    \node at (10.5,4.5){$X_{T-1}$};
       
    \node at (1.7,4.8){$A$};
    \node at (4.2,4.8){$A$};
    \node at (6.7,4.8){$A$};
    \node at (9.2,4.8){$A$};
    
    \node at (0.2,2.1){$B$};
    \node at (2.7,2.1){$B$};
    \node at (5.2,2.1){$B$};
    \node at (10.2,2.1){$B$};
    
     \draw[thick,color=black,->] (1.075,4.5) -- (2.425,4.5);
     \draw[thick,color=black,->] (3.575,4.5) -- (4.925,4.5);
     \draw[thick,color=black,->] (6.075,4.5) -- (7.425,4.5);
     \draw[thick,color=black,->] (8.575,4.5) -- (9.925,4.5);

     \draw[thick,color=black,->] (0.5,3.925) -- (0.5,1);
     \draw[thick,color=black,->] (3.0,3.925) -- (3.0,1);
     \draw[thick,color=black,->] (5.5,3.925) -- (5.5,1);
     \draw[thick,color=black,->] (10.5,3.925) -- (10.5,1);

    \draw[thick,dashed,color=red] (-0.3,3) -- (11.2,3);
   
    \end{tikzpicture}
}
\caption{Hidden Markov Model~\cite{introStamp}}\label{fig:HMM}
\end{figure}
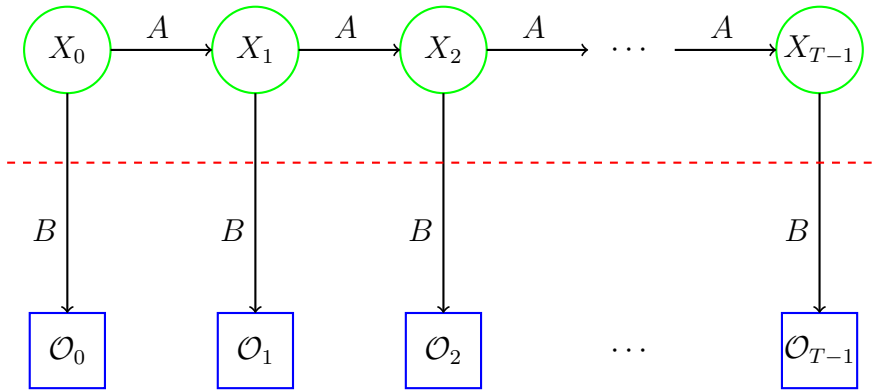

Hidden Markov Models (HMM)~\cite{hmm} can be described as statistical Markov models 
in which the states are hidden. An HMM can be represented by the triple~$\lambda=(A, B, \pi)$, 
where~$A$ is the state transition probability matrix, $B$ is the observation probability matrix, 
and~$\pi$ is the initial state distribution. 
A series of observations, denoted as~$\cO$, are available, and these observations are
probabilistically related to the hidden states sequence~$X$ via the~$B$ matrix.
Figure~\ref{fig:HMM} provides a high-level view of an HMM.

The number of hidden states in an HMM is denoted as~$N$ and the number of unique observation symbols is
denoted as~$M$, while the length of the observation sequence is~$T$. Within the HMM framework, 
there are efficient algorithms to solve three problems~\cite{mshmm}. For the research in this paper, 
we are only focused on the following two problems. 
\begin{enumerate}
\item Given a model~$\lambda = (A, B, \pi)$ and an observation sequence~$\cO$, we can determine 
the optimal hidden state sequence corresponding to~$\cO$, where ``optimal'' is defined as maximizing
the expected number of correct states. Note that this implies an HMM is an Expectation Maximization (EM)
technique. Also, the HMM solution to this problem differs, in general, from a dynamic program, where
we maximize with respect to the overall path.
\item Given an observation sequence~$\cO$ and a specified number of hidden states~$N$, we can
train an HMM. That is, we can determine the matrices that comprise the model~$\lambda = (A, B, \pi)$,
so that~$P(\cO\,|\,\lambda)$ is maximized. 
\end{enumerate}
The so-called forward algorithm and the backward algorithm enable an efficient meet-in-the-middle 
approach to solve problem~1, above~\cite{hmmForwardAndBackward}. 
Typically, the Baum-Welch re-estimation algorithm, which is a hill climb technique,
is used to train an HMM to model a given observation sequence, which solves problem~2, above.

\subsection{Convolutional Neural Network (CNN)}

A convolution can be described as a composite function that computes the amount of overlap of one function as it is 
shifted over another function. In case of discrete sequences~$x$ and~$y$, the convolution is denoted as
\begin{align*}
    c = x*y
\end{align*}
which is computed as
\begin{align*}
    c_{k} = \sum_{i} x_{i}y_{k-i} .
\end{align*}
Here, $c_{k}$ denotes the~$k^{\thth}$ element of the resulting sequence~$c$ and
the summation is performed over all indices~$i$ where the sequence~$x$ and~$y$ overlap. 
The term~$y_{k-i}$ represents the element of sequence~$y$, shifted by~$k$ positions. Note that 
for each position~$k$, the value~$c_{k}$ is computed by summing the products of corresponding 
elements of~$x$ and the shifted version of~$y$.

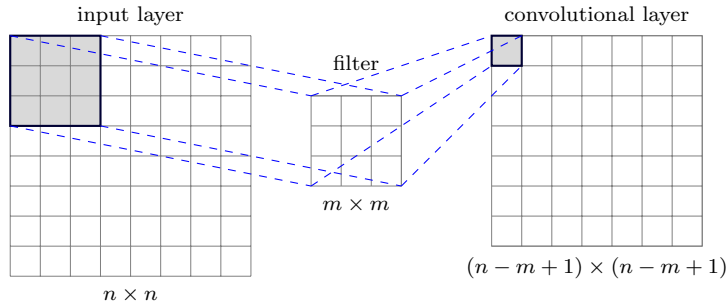
\begin{figure}[!htb]
\centering
\begin{tikzpicture}

    \draw[step=0.4, gray, very thin] (0,0) grid (3.2,3.2);
    \fill[gray!30] (0,3.2) rectangle (1.2,2);
    \draw[step=0.4, gray, very thin] (0,2) grid (1.2,3.2);
    \draw[thick, black!80!blue] (0,3.2) rectangle (1.2,2);
    \node[below, font = \scriptsize] at (1.6, 0) {$n \times n$};
    \node[above, font = \scriptsize] at (1.6, 3.2) {input layer};
    
    \draw[step=0.4, gray, very thin] (4,1.2) grid (5.2, 2.4);
    \draw[gray, very thin] (4,1.2) rectangle ++(1.2,1.2);
    \node[below, font = \scriptsize] at (4.6, 1.2) {$m \times m$};
    \node[above, font = \scriptsize] at (4.6, 2.6) {filter};

    \draw[step=0.4, gray, very thin] (6.4,0.4) grid ++(2.8,2.8);
    \draw[gray, very thin] (6.4,0.4) rectangle ++(2.8,2.8);
    \node[below, font = \scriptsize] at (7.8, 0.4) {$(n-m+1) \times (n-m+1)$};
    \node[above, font = \scriptsize] at (7.8, 3.2) {convolutional layer};

    \coordinate (A) at (0,2);
    \coordinate (B) at (0,3.2);
    \coordinate (C) at (1.2,3.2);
    \coordinate (D) at (1.2,2);
        
    \coordinate (E) at (4,1.2);
    \coordinate (F) at (4,2.4);
    \coordinate (G) at (5.2,2.4);
    \coordinate (H) at (5.2,1.2);
        
    \draw[dashed, blue] (A) -- (E);
    \draw[dashed, blue] (B) -- (F);
    \draw[dashed, blue] (C) -- (G);
    \draw[dashed, blue] (D) -- (H);

    \fill[gray!30] (6.4,3.2) rectangle ++(0.4,-0.4);
    \draw[thick, black!80!blue] (6.4,3.2) rectangle ++(0.4,-0.4);

    \coordinate (A) at (4,1.2);
    \coordinate (B) at (4,2.4);
    \coordinate (C) at (5.2,2.4);
    \coordinate (D) at (5.2,1.2);

    \coordinate (E) at (6.4,3.2);
    \coordinate (F) at (6.8,3.2);
    \coordinate (G) at (6.8,2.8);
    \coordinate (H) at (6.4,2.8);
        
    \draw[dashed, blue] (A) -- (H);
    \draw[dashed, blue] (B) -- (E);
    \draw[dashed, blue] (C) -- (F);
    \draw[dashed, blue] (D) -- (G);
    
\end{tikzpicture}
\caption{Convolution using filter}
\label{fig:conv_layer}
\end{figure}

CNNs~\cite{cnn} are a class of deep neural networks that applies layers of convolution to the input dataset 
using trainable filters. CNNs use a unique architecture to automatically learn and extract hierarchical characteristics 
from data, drawing inspiration from the human visual system. Hence, they are particularly useful for tasks such as feature extraction, 
object detection, and image classification. Figure~\ref{fig:conv_layer} illustrates a convolution applied to input data using a filter.

\begin{figure}[!htb]
\centering
\includegraphics[scale=0.25]{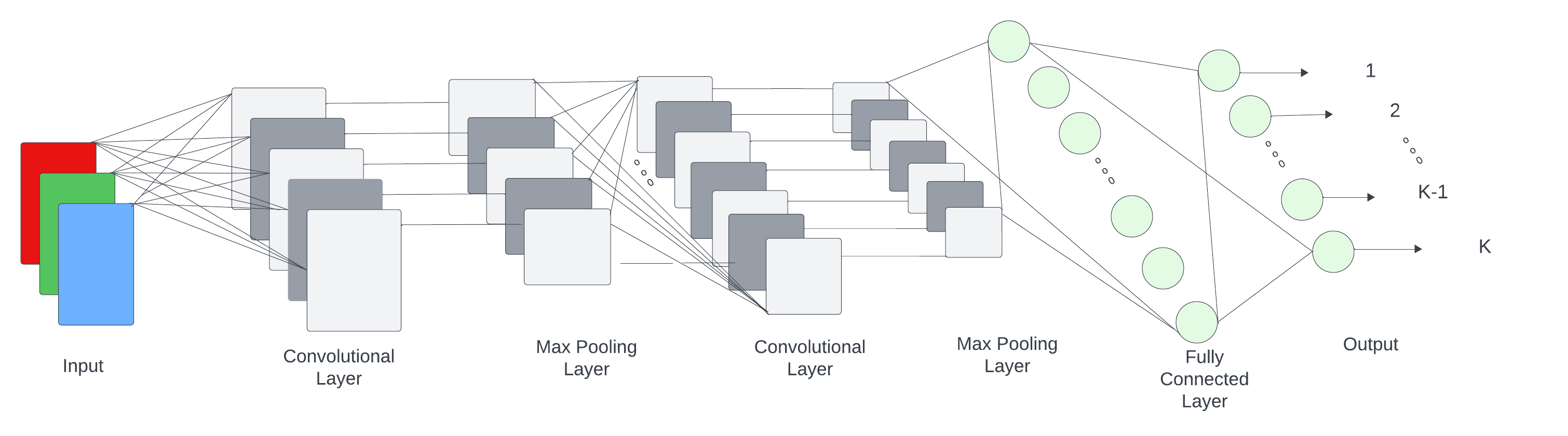}
\caption{Overview of CNN architecture}\label{fig:CNN_architecture}
\end{figure}

In addition to convolutional layers, our CNNs also include max-pooling and fully-connected layers. Max pooling layers 
consist of a non-trainable, fixed filter, which selects the maximum value within non-overlapping windows. Pooling layers
serve primarily to reduce the dimensionality of the data.
Lastly, an activation function is applied via one or more fully connected layers, 
which results in a classification based on the result. Figure~\ref{fig:CNN_architecture} 
illustrates a generic CNN architecture.

\section{Literature Review}\label{sect:review}

There has been a vast amount of previous work on malware classification using 
a wide range of machine learning  and deep learning approaches. 
This section discusses a representative sample 
of such malware classifications techniques, with the emphasis on
research that is most closely related to our novel NLP-inspired HMM-CNN technique.

\subsection{Malware Classification using HMM}

In one of the earliest papers in this genre, Wong and Stamp~\cite{Wing} consider HMMs for the detection
of metamorphic malware. By modern standards, they considered a very small sample set, but they
were able to distinguish malware from benign with high accuracy, clearly indicating the viability of
machine learning models within the malware domain.

Annachhatre et al.~\cite{hmmmalwareclass} train multiple HMMs on a variety of metamorphic malware samples.
Each malware sample in the test set is then scored against all models, and the samples are clustered
based on the resulting vector of scores.
They were able to classify the malware samples into their respective families with good accuracy,
even for families that were not included in the training set.

In~\cite{gmmhmm}, Zhao et al., explore the usage of complex Gaussian Mixture Model-HMMs (GMM-HMM) 
for malware classification. In their research, GMM-HMMs produced comparable results to discrete HMMs 
based on opcode sequence features, and showed significant improvement over discrete HMMs when trained on 
entropy-based features.

\subsection{Malware Classification using SVM}

Support Vector Machines (SVM) are a prominent class of techniques for supervised learning. 
The objective of the SVM algorithm is to determine an optimal hyperplane---or hyperplanes, in the
the more general multiclass case---that can segregate $n$-dimensional space 
into classes. The decision boundary is then used to classify data points not in the training set. 
In~\cite{svm1}, Kruczkowski et al., trained an SVM on malware samples and achieved a 
cross-validation accuracy of~0.9398, and an F1-score of~0.9552. 

Singh et al.~\cite{svm2} also use SVMs for malware classification. They trained 
HMMs, computed a Simple Substitution Distance (SSD) score based on the classic encryption 
technique from symmetric cryptography, and also computed an Opcode Graph Score (OGS). 
Each malware sample was classified---using an SVM---based on its vector of these three scores. 
While the individual scores generally performing poorly in a robustness analysis, 
the SVM results were significantly more robust, indicating the advantage of combining 
multiple scores via an SVM.

\subsection{Malware Classification using Random Forest}

In~\cite{randomforest1}, Garcia and Muga~II employ an approach for converting a binary file to a gray scale image, 
and subsequently use a Random Forest to classify malware into families, achieving an accuracy of~0.9562.
Domenick et al.~\cite{randomforest2}, on the other hand, combine a Random Forest with 
Principal Component Analysis (PCA)~\cite{pca} and Term Frequency-Inverse Document Frequency (TF-IDF)~\cite{tfidf}.
The model based on a Random Forest and PCA outperformed a models based on Logistic Regression, 
Decision Trees, and SVM on a particular dataset.

\subsection{Malware Classification using RNN and LSTM}

A Recurrent Neural Network (RNN)~\cite{rnnRef} is a type of neural network designed 
to process sequential data by incorporating feedback connections. This gives RNNs
a form of memory that is lacking in feedforward neural networks.
However, generic RNNs are subject to computational issues, including
vanishing and exploding gradients, which limit their utility. Consequently, various specialized RNN-based
architectures have been developed, which mitigate some of the issues observed in plain vanilla RNNs.
The best-known and most successful of these specialized RNN architectures is the 
Long Short-Term Memory (LSTM) model.

An unsupervised approach involving Echo State Networks (ESNs)~\cite{esns} and RNNs for a 
``projection'' stage to extract features is discussed by Pascanu et al.~\cite{malwareclassrnn}. 
A standard classifier then uses these extracted features to detect malicious samples. Their hybrid model 
with the best performance employed ESN for the recurrent model, a max pooling layer 
for nonlinear sampling, and Logistic Regression for the final classification.

R.~Lu~\cite{lstmmalwareclass}, experimented with LSTMs for malware classification. 
First, Word2Vec word embedding of the opcodes were generated using skip-gram and CBOW models. 
Subsequently, a two stage LSTM model was used for malware detection. The two-stage LSTM model 
is composed of two LSTM layers and one mean-pooling layer to obtain feature representations of 
malware opcode sequences. An average Area under the ROC Curve (AUC)~\cite{auc} of~0.987 was achieved for malware 
classification on a modest-sized dataset consisting of~969 malware and~123 benign samples.

\subsection{Malware Classification using CNN}

Recently, image-based analysis of malware has been the focus of considerable
research; see~\cite{Niket,Mugdha,Huy,Sravani}, for examples. Much of this work is based
on CNNs~\cite{cnn}. A CNN is a type of neural network that designed
to efficiently deal with data that is in a grid-like layout where local structure dominates, which 
is the case for images. In~\cite{cnn-mal-class}, Kalash et al., proposed a CNN-based architecture, 
called M-CNN, for malware classification. The architecture of M-CNN is based on the 
VGG-16~\cite{vgg16}, and it achieves accuracies of~0.9852 and~0.9997 on the 
popular MalImg~\cite{malimg} dataset and a Microsoft~\cite{microsoft-dataset} dataset, 
respectively.

\section{Methodology}\label{sect:meth}

In this section, we first introduce the dataset used in our experiments. We then outline
the experimental design that we employ for the results presented in Section~\ref{sect:exp}.

\subsection{Dataset and Preprocessing}\label{sect:DP}

As in~\cite{hmm-rf}, for the research presented here, we use the malware samples in the
popular Malicia dataset~\cite{Malicia}. 
This dataset includes~11,688 malware binaries, categorized into~48 different malware families. The binary 
files were gathered from~500 drive-by download servers over a period of~11 months. They were then executed 
in a virtualized environment designed to capture the network traffic produced by the malware and to take a screenshot 
of the guest VM at the end of the execution. Windows XP Service Pack~3 was used as the guest operating system. 
To classify the binaries into malware families, a combination of automatic clustering techniques and an analyst that 
manually refines the generic labels by comparing cluster behaviors against public reports were employed. 

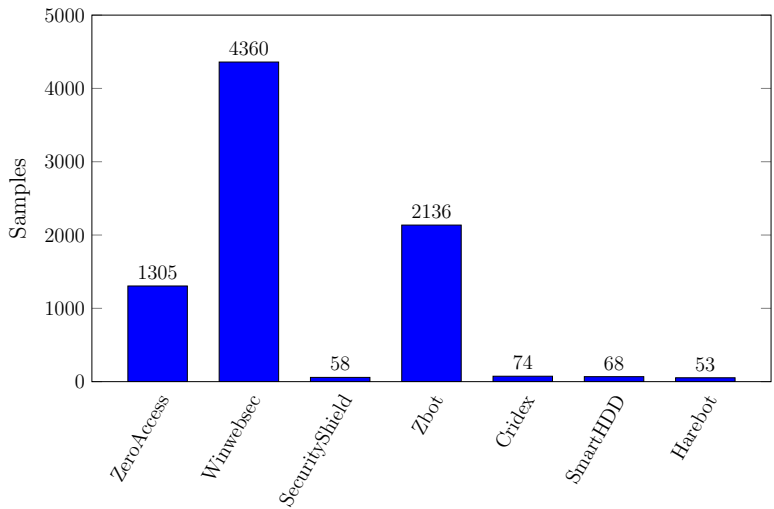
\begin{figure}[!htb]
\centering
\begin{tikzpicture}[scale=0.9, every node/.style={scale=0.9}]
\pgfkeys{/pgf/number format/.cd,1000 sep={}}
\begin{axis}[
        width  = 0.75*\textwidth,
        height = 7.0cm,
        ymin=0,ymax=5000,
        ytick={0,1000,2000,3000,4000,5000},
        major x tick style = transparent,
        ybar=5*\pgflinewidth,
        bar width=25.0pt,
        ylabel = {Samples},
        symbolic x coords={ZeroAccess,Winwebsec,SecurityShield,Zbot,Cridex,SmartHDD,Harebot},
        xticklabels={ZeroAccess,Winwebsec,SecurityShield,Zbot,Cridex,SmartHDD,Harebot},
        label style={scale=0.90},
	y tick label style={
		scale=0.80,
    		/pgf/number format/.cd,
   		fixed,
   		fixed zerofill,
    		precision=0},
        xtick = data,
        x tick label style={
        		rotate=60,
		scale=0.80,
		anchor=north east,
		inner sep=0mm
		},
        nodes near coords,
        every node near coord/.append style={
								   scale=0.80,
								   /pgf/number format/.cd,
								   fixed,
								   fixed zerofill,
								   precision=0},
        enlarge x limits=0.12,
        legend cell align=left,
        legend style={
                at={(0.91,0.02)},
                anchor=south,
                column sep=1ex
        },
]
\addplot [fill=blue,opacity=1.00]
coordinates {
(ZeroAccess, 1305)
(Winwebsec, 4360)
(SecurityShield, 58)
(Zbot, 2136)
(Cridex, 74)
(SmartHDD, 68)
(Harebot, 53)
};
\end{axis}
\end{tikzpicture}
\caption{Malware samples per family}\label{fig:Dataset distribution}
\end{figure}

The Malicia dataset is highly imbalanced and hence we remove all classes with less than~50 samples. 
This results in malware samples belonging to the following seven families.
\begin{description}
    \item\textbf{Zeroaccess} tries to steal information, and it can also cause other malicious actions, 
    such as downloading additional malware or opening a backdoor~\cite{zeroaccess}.
    \item \textbf{Winwebsec} is a Trojan horse that attempts to install additional malicious programs~\cite{winwebsec}.
    \item \textbf{SecurityShield} is based on Winwebsec, and it displays fake security warnings in an attempt to get the user to 
    pay money to fix the nonexistent issues~\cite{securityShield}.
    \item \textbf{Zbot} is a Trojan that tries to steal user information. It spreads by attaching executable files to spam email messages~\cite{zbot}.
    \item \textbf{Cridex} is a worm that installs a backdoor that can then be used to download additional malware onto a system~\cite{cridex}.
    \item \textbf{SmartHDD} pretends to be a hard drive optimizer. SmartHDD finds multiple nonexistent issues, and attempts to 
    convince the user to pay money to ``repair'' the hard drive~\cite{smartHDD}.
    \item \textbf{Harebot} is a rootkit that opens a system to remote attacks of various types~\cite{Harebot}.
\end{description}
The number of samples in each of these malware families is shown in Figure~\ref{fig:Dataset distribution}.

\subsection{Experimental Design}

The first step in our experimental design is to disassemble every executable file in the dataset 
and extract mnemonic opcode sequences. Next, the dataset is split into train and test sets. 
For all of our experiments, we use an~\hbox{80:20} train-test split, i.e., 80\%\ of the samples 
are used for training, while 20\%\ of the samples are reserved for testing. We trained the models 
used in this research on a PC, with the specification of this machine 
shown in Table~\ref{tab:hardsoft}; the software---including operating system, 
and Python packages used---is also specified in Table~\ref{tab:hardsoft}.

\begin{table}[!htb]
\centering
\caption{Relevant hardware and software}\label{tab:hardsoft}
\adjustbox{scale=0.85}{
\begin{tabular}{c|cc}\midrule\midrule
 & Item & Version\\ \midrule
\multirow{4}{*}{Hardware}
& Chip & Apple~M1 Pro\\
& Cores & 8\\
& Memory & 16~GB\\
& Firmware Version & 8422.121.1\\ \midrule
\multirow{6}{*}{Software}
& OS & macOS Ventura\\
& Python & 3.9.12\\
& NumPy & 1.21.5\\
& Pandas & 1.4.2\\
& Pickle & 4.0\\
& Scikit Learn & 1.0.2\\
\midrule\midrule
\end{tabular}
}
\end{table}

\subsection{Training Methodology}\label{subsec: training_procedure}

The methodology for training our HMM-CNN model can be summarized in the following six steps. 
\begin{enumerate}
    \item Train HMMs on opcode sequences --- This step consists of training seven different HMMs;
    one HMM for each malware family discussed in Section~\ref{sect:DP}, above. 
    Each HMM is trained using only the opcode sequences of samples belonging to a particular family. 
    The observation sequence~$\cO$  for a given malware family is obtained 
    by concatenating the observation sequences (i.e., mnemonic opcode sequences) extracted from 
    training samples belonging to the family. 
    When training these HMMs, we specify the number of hidden states~$N$, which is a hyperparameter
    of our overall system. We experiment with different choices for~$N$.
    \item Determine the feature vector for each sample --- The first~$L$ opcodes 
    of a given sample are fed into each of the seven trained HMMs. 
    This results in seven hidden state sequence vectors that are each of length~$L$.
    We concatenate these seven hidden state sequences to obtain a feature vector of length~$7L$. 
    The length~$L$ is a hyperparameter of the system, and hence we experiment with different
    choices for~$L$.
    \item Scale the feature vectors --- Each feature vector is scaled using a standard scaler, that is, element~$x$ is
    scaled as~$z=(x-\mu)/\sigma$, where~$\mu$ is the mean and~$\sigma$ is the standard deviation.
    \item Generate images from feature vectors --- Each scaled hidden state sequence obtained in the previous step 
    is formed into a square matrix. Since each vector is of length~$7L$ 
    these square matrices are of size~$\lceil\sqrt{7L}\rceil\times\lceil\sqrt{7L}\rceil$.
    These matrices are then padded with zeros at the edges to create images of dimension~$224\times 224$. 
    The padding is applied evenly across all the edges to maintain the symmetry.
    \item Select the CNN architecture --- In this step, a popular CNN architecture is chosen as a base model for 
    our HMM-CNN architecture. We then add custom classification layers (as discussed below)
    on top of our base model so that the resulting CNN can classify images into the seven malware families. 
    We treat the CNN architecture as a hyperparameter of our overall model, and hence we experiment 
    with several different base models.
    \item Train the CNN model --- Lastly, we train the CNN model on the images discussed above. 
    Of course, the malware family to which an image belongs serves as its label.
\end{enumerate}
To summarize, we train an HMM for each family, then use the trained HMMs to determine the hidden state sequences 
corresponding to each sample. We rearrange these hidden state sequence vectors to square matrices, then
form images of size~$224\times 224$, which are used to train a CNN. Each CNN architecture is built upon a base model 
that includes the following three additional layers.
\begin{enumerate}
\item A Global Average Pooling (GAP) layer~\cite{pooling_layer} is used to reduce
spatial dimensions. 
\item The GAP layer is followed by a dense layer with~1024 neurons and ReLU activation~\cite{relu}, 
which serves as a ``bottleneck'', in the sense that it forces the model to condense the 
most relevant features into a more compact representation, which serves to reduce overfitting. 
\item The final dense layer, with softmax activation~\cite{softmax}, has its number of neurons equal 
to the number of malware families and is responsible for classifying input data. In all of our experiments,
the number of classes is seven, since we consider seven malware families from the Malicia dataset.
\end{enumerate}

As mentioned above, we experiment with several base CNN architectures. 
Next, we provide a brief description of each of the base CNN architectures that we consider.

\begin{description}
    \item \textbf{ResNet50V2} is part of the Residual Network (ResNet) family.
    As indicated by its name, ResNet50V2 contains~50 layers and is noteworthy for its 
    deep architecture and skip connections, allowing it to excel at image classification tasks~\cite{resnet}.
    \item \textbf{ResNet101V2} is an extended version of the ResNet architecture with~101 layers. 
    Similar to other ResNet models, it has skip connections to facilitate the training of extremely deep networks~\cite{resnet}.
    \item \textbf{ResNet152V2} is another variant of the ResNet architecture, in this case with~152 layers. 
    ResNet152V2 addresses the vanishing gradient problem by using residual connections~\cite{resnet}.
    \item \textbf{DenseNet201} is a deep neural network model that uses dense connections between layers. 
    This model has~201 layers and is known for efficient feature reuse~\cite{densenet}.
    \item \textbf{Xception} is known for its extreme depth and parallelism. It employs depthwise separable convolutions, 
    making it computationally efficient while achieving high performance in image classification tasks~\cite{xception}.
\end{description}

Table~\ref{tab: HMMCNN_param_desc} provides a brief summary
of each of the hyperparameters of our HMM-CNN model.
Recall that the resulting feature vectors are of length~$7L$, and images 
of  size~$224\times 224$ are generated from these feature vectors.

\begin{table}[!htb]
\centering
\caption{HMM-CNN hyperparameters}\label{tab: HMMCNN_param_desc}
\adjustbox{scale=0.85}{
\begin{tabular}{c|c}\midrule\midrule
Hyperparameter & Description\\ \midrule
$N$ & Number of hidden states in the HMM\\
$L$ & Length of each extracted hidden state sequence\\
\texttt{base\_model} & Base CNN architecture\\
\texttt{optimizer} & Algorithm to adjust model parameter during training\\
\texttt{learning\_rate} & Step size at which the model parameter are updated\\ 
\texttt{loss} & Quantifies difference between actual and predicted output\\ \midrule\midrule
\end{tabular}
}
\end{table}

\section{Experiments and Results}\label{sect:exp}

In this section, we first discuss the HMM training and the use to the resulting models to obtain
hidden state sequences. Next, we consider the training of our HMM-CNN classifier,
including hyperparameter tuning.
Then we summarize the results of our experiments, and we conclude this section with a comparison of 
our results to other research involving the Malicia dataset.

\subsection{HMM Training and Hidden States}

As discussed above, the subset of the Malicia dataset that we use
consists of seven malware families, and we train one HMM for each family.
Hence, we have seven trained HMMs, where each model is of the form~$\lambda=(A, B, \pi)$.
We experimented with the number of hidden states~$N\in\{5,10,20,30\}$,
and we found that~$N=20$ yields the highest accuracy. The number of unique observations 
(i.e., a superset of the opcodes in all seven families) is~$426$, with \texttt{MOV} being the most frequent. 
Therefore, $N=20$ and~$M = 426$ in all of our HMMs
discussed in the remainder of this paper. 

Recall that these HMM matrices are~$A=\{a_{ij}\}$, which is~$N\times N$,
$B=\{b_{ij}\}$, which is~$N\times M$, and~$\pi=\{\pi_i\}$, which is~$1\times N$.
We initialize the~$A$, $B$, and~$\pi$ matrices to approximately uniform, that is, 
each~$a_{ij} \approx 1/N$, each~$b_{ij} \approx 1/M$, and each~$\pi_i \approx 1/N$,
while enforcing the required row stochastic conditions.
The minimum number of iterations of the Baum-Welch re-estimation algorithm
is set to~10, and we stop when successive iterations beyond this number produce a 
change in~$P(\cO\,|\,\lambda)$ of less than~$\varepsilon = 0.001$. 
When training our models, the average number of iterations was~10.43, 
and it took an average of five hours to train each HMM. Note that this is one-time
work.

Next, we use the trained HMMs to generate hidden state sequences for each sample as follows.
Given a sample, we generate a hidden state sequence using 
each of the seven HMMs.
The length of each hidden state sequence 
corresponding to each malware sample is truncated to a constant~$L$, that is, we only use the 
hidden state sequences corresponding to the first~$L$ opcodes. We experiment with~$L\in\{25,50,100,200\}$.
In rare cases, there were insufficient opcodes available in a given sample, 
(i.e., the length of opcode sequence for the malware sample was less than~$L$), 
in which case we dropped
the sample from consideration. The number of dropped samples for
each value of~$L$ is given in Table~\ref{tab: malware_sample_dropped},
and we observe that an insignificant
percentage of malware samples were dropped.

\begin{table}[!htb]
\centering
\caption{Number of malware sample dropped for different values of $L$}
\label{tab: malware_sample_dropped}
\adjustbox{scale=0.85}{%
\begin{tabular}{c|c}\midrule\midrule
$L$ & Samples dropped\\ \midrule
\zz25\zz & \zz3 \\
\zz50\zz & 11 \\
100\zz & 14 \\
200\zz & 26 \\ \midrule\midrule
\end{tabular}
}
\end{table}

\subsection{HMM-CNN Training}

In our proposed HMM-CNN technique, we train a CNN on images created by reshaping the hidden state sequences 
generated by HMMs. As discussed above, for each sample, the concatenated hidden state sequence vector of length~$7L$ 
is rearranged to a square matrix, and this matrix is then padded with zeros at its edges to create an image of 
size~$224\times 224$. Note that this image dimension was chosen because most modern CNN architectures 
cannot be trained on images of dimensions smaller than~$224\times 224$. Also, we found that this embedding
approach yielded slightly better results than resizing the images.

We conduct a grid-search~\cite{grid-search} 
to determine the hyperparameters of our HMM-CNN classifier. 
Specifically, we tested the hyperparameter values in Table~\ref{tab: param_HMMCNN}, 
with the values in boldface yielding the best result.
The accuracy obtained for the best choice of hyperparameters in 
Table~\ref{tab: param_HMMCNN} was~$0.9781$. 

\begin{table}[!htb]
\centering
\caption{HMM-CNN hyperparameters tested and selected}\label{tab: param_HMMCNN}
\adjustbox{scale=0.8}{
\begin{tabular}{c|c}\midrule\midrule
Hyperparameter & Tested (selected in boldface)\\ \midrule
$L$ & 56, \textbf{112}, 224 \\
\texttt{base\_model} & \textbf{ResNet50V2}, ResNet152V2, ResNet101V2, DenseNet201, Xception\\
\texttt{optimizer} & Adam, RMSProp, Adagrad, Adadelta, \textbf{Nadam}, Ftrl\\
\texttt{learning\_rate} & 0.0001, \textbf{0.001}, 0.01\\
\texttt{loss} & categorical\_crossentropy, \textbf{kullback\_leibler\_divergence}, poisson \\ \midrule\midrule
\end{tabular}
}
\end{table}

We give expanded results for each of the individual hyperparameters of HMM-CNN in Figure~\ref{fig:hmmcnnGraph_Features}. 
We observe that a \texttt{learning\_rate} of~0.001 and the Categorical Crossentropy \texttt{loss} function were both clearly
superior to the alternatives that we tested. For the choice of \texttt{optimizer}, the results are not as clear, but Nadam was
generally the best.

\begin{figure}[!htb]
\centering
\begin{tabular}{cc}
\multicolumn{2}{c}{\begin{tikzpicture}[scale=0.65, every node/.style={scale=1.0}]
\pgfkeys{/pgf/number format/.cd,1000 sep={}}
\begin{axis}[
        width  = 1.125*\textwidth,
        height = 7.5cm,
        ymin=0.60,ymax=1.0,
        ytick={0.60, 0.65, 0.70, 0.75, 0.80, 0.85, 0.90, 0.95, 1.00},
        major x tick style = transparent,
        ybar=5*\pgflinewidth,
        bar width=9.5pt,
        ylabel = {Accuracy},
        symbolic x coords={Adam, RMSProp, Adagrad, Adadelta, Nadam, Ftrl},
        xticklabels={Adam, RMSProp, Adagrad, Adadelta, Nadam, Ftrl},
	y tick label style={
    		/pgf/number format/.cd,
   		fixed,
   		fixed zerofill,
    		precision=2},
        xtick = data,
        x tick label style={
		},
        enlarge x limits=0.1,
        legend cell align=left,
        legend style={
                at={(0.9,0.57)},
                anchor=south,
                column sep=1ex
        },
]
\addplot [fill=blue,opacity=1.00]
coordinates {
(Adam, 0.8157)
(RMSProp, 0.8421)
(Adagrad, 0.8954)
(Adadelta, 0.8567)
(Nadam, 0.8788)
(Ftrl, 0.7492)
};
\addlegendentry{ResNet50V2}
\addplot [fill=red,opacity=1.00]
coordinates {
(Adam, 0.8446)
(RMSProp, 0.8294)
(Adagrad, 0.8663)
(Adadelta, 0.7592)
(Nadam, 0.8737)
(Ftrl, 0.7516)
};
\addlegendentry{DenseNet201}
\addplot [fill=yellow,opacity=1.00]
coordinates {
(Adam, 0.8126)
(RMSProp, 0.8672)
(Adagrad, 0.8151)
(Adadelta, 0.7054)
(Nadam, 0.9216)
(Ftrl, 0.6319)
};
\addlegendentry{Xception}
\addplot [fill=black,opacity=1.00]
coordinates {
(Adam, 0.8887)
(RMSProp, 0.8031)
(Adagrad, 0.8817)
(Adadelta, 0.7825)
(Nadam, 0.9732)
(Ftrl, 0.6896)
};
\addlegendentry{ResNet152V2}
\addplot [fill=green,opacity=1.00]
coordinates {
(Adam, 0.8162)
(RMSProp, 0.8430)
(Adagrad, 0.9080)
(Adadelta, 0.8840)
(Nadam, 0.8145)
(Ftrl, 0.7579)
};
\addlegendentry{ResNet101V2}
\end{axis}
\end{tikzpicture}} 
\\
\multicolumn{2}{c}{\adjustbox{scale=0.85}{(a) \texttt{Optimizer}}}
\\
\\
\begin{tikzpicture}[scale=0.65, every node/.style={scale=1.0}]
\pgfkeys{/pgf/number format/.cd,1000 sep={}}
\begin{axis}[
        width  = 0.65*\textwidth,
        height = 6.75cm,
        ymin=0.68,ymax=0.96,
        ytick={0.68, 0.72, 0.76, 0.80, 0.84, 0.88, 0.92, 0.96},
        major x tick style = transparent,
        ybar=5*\pgflinewidth,
        bar width=8.0pt,
        ylabel = {Accuracy},
        symbolic x coords={0.0001, 0.001, 0.01},
        xticklabels={0.0001, 0.001, 0.01},
	y tick label style={
    		/pgf/number format/.cd,
   		fixed,
   		fixed zerofill,
    		precision=2},
        xtick = data,
        x tick label style={
		},
        enlarge x limits=0.3,
        legend cell align=left,
        legend style={
                at={(0.85,0.5)},
                anchor=south,
                column sep=1ex
        },
]
\addplot [fill=blue,opacity=1.00]
coordinates {
(0.0001, 0.8555)
(0.001, 0.9192)
(0.01, 0.7340)
};
\addlegendentry{ResNet50V2}
\addplot [fill=red,opacity=1.00]
coordinates {
(0.0001, 0.8371)
(0.001, 0.9169)
(0.01, 0.7454)
};
\addlegendentry{DenseNet201}
\addplot [fill=yellow,opacity=1.00]
coordinates {
(0.0001, 0.8192)
(0.001, 0.8593)
(0.01, 0.7505)
};
\addlegendentry{Xception}
\addplot [fill=black,opacity=1.00]
coordinates {
(0.0001, 0.8470)
(0.001, 0.8855)
(0.01, 0.8093)
};
\addlegendentry{ResNet152V2}
\addplot [fill=green,opacity=1.00]
coordinates {
(0.0001, 0.8581)
(0.001, 0.9270)
(0.01, 0.6986)
};
\addlegendentry{ResNet101V2}
\end{axis}
\end{tikzpicture}
&
\begin{tikzpicture}[scale=0.65, every node/.style={scale=1.0}]
\pgfkeys{/pgf/number format/.cd,1000 sep={}}
\begin{axis}[
        width  = 0.65*\textwidth,
        height = 6.75cm,
        ymin=0.70,ymax=0.94,
        ytick={0.70, 0.74, 0.78, 0.82, 0.86, 0.90, 0.94},
        major x tick style = transparent,
        ybar=5*\pgflinewidth,
        bar width=8.0pt,
        ylabel = {Accuracy},
        symbolic x coords={categorical_crossentropy, kl_divergence, poisson},
        xticklabels={Crossentropy, KL Divergence, Poisson},
	y tick label style={
    		/pgf/number format/.cd,
   		fixed,
   		fixed zerofill,
    		precision=2},
        xtick = data,
        x tick label style={
		},
        enlarge x limits=0.3,
        legend cell align=left,
        legend style={
                at={(0.85,0.5)},
                anchor=south,
                column sep=1ex
        },
]
\addplot [fill=blue,opacity=1.00]
coordinates {
(categorical_crossentropy, 0.9206)
(kl_divergence, 0.8216)
(poisson, 0.7704)
};
\addlegendentry{ResNet50V2}
\addplot [fill=red,opacity=1.00]
coordinates {
(categorical_crossentropy, 0.9107)
(kl_divergence, 0.8350)
(poisson, 0.7399)
};
\addlegendentry{DenseNet201}
\addplot [fill=yellow,opacity=1.00]
coordinates {
(categorical_crossentropy, 0.8749)
(kl_divergence, 0.7879)
(poisson, 0.7466)
};
\addlegendentry{Xception}
\addplot [fill=black,opacity=1.00]
coordinates {
(categorical_crossentropy, 0.9045)
(kl_divergence, 0.8261)
(poisson, 0.7991)
};
\addlegendentry{ResNet152V2}
\addplot [fill=green,opacity=1.00]
coordinates {
(categorical_crossentropy, 0.9243)
(kl_divergence, 0.7949)
(poisson, 0.7751)
};
\addlegendentry{ResNet101V2}
\end{axis}
\end{tikzpicture}
\\
(b) \adjustbox{scale=0.85}{\texttt{Learning rate}}
&
(c) \adjustbox{scale=0.85}{\texttt{Loss function}}
\end{tabular}
\caption{Accuracy trends for different hyperparameters for HMM-CNN}\label{fig:hmmcnnGraph_Features}
\end{figure}

\begin{figure}[!htb]
\centering
\begin{tabular}{cc}
\begin{tikzpicture}[scale=0.45]
    \begin{axis}[
        width=10cm,
        height=10cm,
	colormap={bluewhite}{color=(white) rgb255=(100,149,237)},
        xticklabels={ZeroAccess,Winwebsec,SecurityShield,Zbot,Cridex,SmartHDD,Harebot},
        xtick={0,...,6},
        xtick style={draw=none},
	xticklabel style={anchor=east,rotate=45,yshift=-5pt,font=\large},
        yticklabels={ZeroAccess,Winwebsec,SecurityShield,Zbot,Cridex,SmartHDD,Harebot},
        ytick={0,...,6},
        ytick style={draw=none},
        enlargelimits=false,
        yticklabel style={font=\large},
        colorbar,
        colorbar style={
            ytick={0,40,80,120,160,200},
            yticklabels={0,40,80,120,160,200},
            yticklabel={\pgfmathprintnumber\tick},
            yticklabel style={
            		scale=1.25,
            		/pgf/number format/fixed,
			/pgf/number format/precision=0}
        },
        point meta min=0,
        point meta max=200,
        nodes near coords={\pgfmathprintnumber\pgfplotspointmeta},
        nodes near coords black white/.style={
            small value/.style={
                yshift=-7pt,
                text=black,
                /pgf/number format/fixed,
                /pgf/number format/precision=0,
                /pgf/number format/zerofill=true,
                scale=1.25,
            },
            large value/.style={
                yshift=-7pt,
                text=white,
                /pgf/number format/fixed,
                /pgf/number format/precision=0,
                /pgf/number format/zerofill=true,
                scale=1.25,
            },
            every node near coord/.style={
                check for zero/.code={
                    \pgfmathfloatifflags{\pgfplotspointmeta}{0}{
                        \pgfkeys{/tikz/coordinate}
                    }{
                        \begingroup
                        \pgfkeys{/pgf/fpu}
                        \pgfmathparse{\pgfplotspointmeta<#1}
                        \global\let\result=\pgfmathresult
                        \endgroup
                        %
                        %
                        \pgfmathfloatcreate{1}{1.0}{0}
                        \let\ONE=\pgfmathresult
                        \ifx\result\ONE
                            \pgfkeysalso{/pgfplots/small value}
                        \else
                            \pgfkeysalso{/pgfplots/large value}
                        \fi
                    }
                },
                check for zero,
            },
        },
        nodes near coords black white=100,
    ]
        \addplot[
            matrix plot,
            mesh/cols=7,
            point meta=explicit,draw=gray
        ] table [meta=C] {
            x y C
0 0 260
1 0 0
2 0 1
3 0 0
4 0 0
5 0 0
6 0 0
0 1 1
1 1 864
2 1 0
3 1 0
4 1 1
5 1 0
6 1 0
0 2 1
1 2 0
2 2 11
3 2 0
4 2 0
5 2 0
6 2 0
0 3 7
1 3 9
2 3 0
3 3 414
4 3 0
5 3 0
6 3 0
0 4 1
1 4 10
2 4 0
3 4 1
4 4 3
5 4 0
6 4 0
0 5 0
1 5 0
2 5 0
3 5 0
4 5 0
5 5 13
6 5 0
0 6 1
1 6 1
2 6 0
3 6 1
4 6 0
5 6 0
6 6 8
         };
    \end{axis}
\end{tikzpicture}
&
\begin{tikzpicture}[scale=0.45]
    \begin{axis}[
        width=10cm,
        height=10cm,
	colormap={bluewhite}{color=(white) rgb255=(100,149,237)},
        xticklabels={ZeroAccess,Winwebsec,SecurityShield,Zbot,Cridex,SmartHDD,Harebot},
        xtick={0,...,6},
        xtick style={draw=none},
	xticklabel style={anchor=east,rotate=45,yshift=-5pt,font=\large},
        yticklabels={ZeroAccess,Winwebsec,SecurityShield,Zbot,Cridex,SmartHDD,Harebot},
        ytick={0,...,6},
        ytick style={draw=none},
        enlargelimits=false,
        yticklabel style={font=\large},
        colorbar,
        colorbar style={
            ytick={0,0.2,0.4,0.6,0.8,1.0},
            yticklabels={0,0.2,0.4,0.6,0.8,1.0},
            yticklabel={\pgfmathprintnumber\tick},
            yticklabel style={
            		scale=1.25,
            		/pgf/number format/fixed,
			/pgf/number format/precision=2}
        },
        point meta min=0.0,
        point meta max=1.0,
        nodes near coords={\pgfmathprintnumber\pgfplotspointmeta},
        nodes near coords black white/.style={
            small value/.style={
                yshift=-7pt,
                text=black,
                /pgf/number format/fixed,
                /pgf/number format/precision=3,
                /pgf/number format/zerofill=true,
                scale=1.1,
            },
            large value/.style={
                yshift=-7pt,
                text=white,
                /pgf/number format/fixed,
                /pgf/number format/precision=3,
                /pgf/number format/zerofill=true,
                scale=1.1,
            },
            every node near coord/.style={
                check for zero/.code={
                    \pgfmathfloatifflags{\pgfplotspointmeta}{0}{
                        \pgfkeys{/tikz/coordinate}
                    }{
                        \begingroup
                        \pgfkeys{/pgf/fpu}
                        \pgfmathparse{\pgfplotspointmeta<#1}
                        \global\let\result=\pgfmathresult
                        \endgroup
                        %
                        %
                        \pgfmathfloatcreate{1}{1.0}{0}
                        \let\ONE=\pgfmathresult
                        \ifx\result\ONE
                            \pgfkeysalso{/pgfplots/small value}
                        \else
                            \pgfkeysalso{/pgfplots/large value}
                        \fi
                    }
                },
                check for zero,
            },
        },
        nodes near coords black white=0.5,
    ]
        \addplot[
            matrix plot,
            mesh/cols=7,
            point meta=explicit,draw=gray
        ] table [meta=C] {
            x y C
0 0 0.9962
1 0 0
2 0 0.0038
3 0 0
4 0 0
5 0 0
6 0 0
0 1 0.0012
1 1 0.9977
2 1 0
3 1 0
4 1 0.0012
5 1 0
6 1 0
0 2 0.0833
1 2 0
2 2 0.9167
3 2 0
4 2 0
5 2 0
6 2 0
0 3 0.0163
1 3 0.0209
2 3 0
3 3 0.9628
4 3 0
5 3 0
6 3 0
0 4 0.0667
1 4 0.6667
2 4 0
3 4 0.0667
4 4 0.2000
5 4 0
6 4 0
0 5 0
1 5 0
2 5 0
3 5 0
4 5 0
5 5 1.0000
6 5 0
0 6 0.0909
1 6 0.0909
2 6 0
3 6 0.0909
4 6 0
5 6 0
6 6 0.7273
         };
    \end{axis}
\end{tikzpicture}
\\[-2ex]
\adjustbox{scale=0.85}{(a) Actual}
&
\adjustbox{scale=0.85}{(b) Scaled}
\end{tabular}
\caption{Confusion matrices for HMM-CNN model}\label{fig:conf_HMM-CNN}
\end{figure}
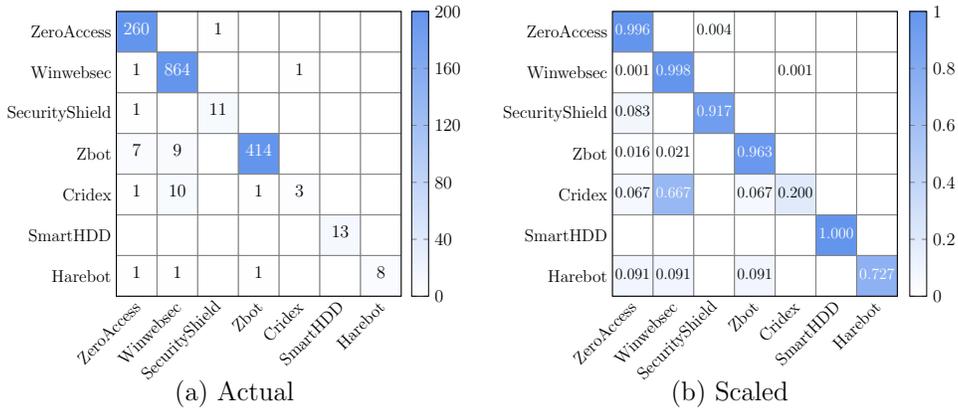

Confusion matrices for our HMM-CNN experimental results are given in Figure~\ref{fig:conf_HMM-CNN}, 
where Figure~\ref{fig:conf_HMM-CNN}(a) provides the actual number of classifications for each case, 
and Figure~\ref{fig:conf_HMM-CNN}(b) is a scaled confusion matrix. The samples belonging to the three largest 
malware families, namely, ZeroAccess, Winwebsec, and Zbot are classified with an average accuracy of~$0.9856$, 
whereas Cridex, with only~74 samples available, is classified with the lowest accuracy of~$0.200$.

\subsection{Comparison to Related Techniques}

Finally, we compared the results obtained from our HMM-CNN model with a variety of related techniques. 
The following provides a brief description of each of the related techniques that we consider.
\begin{itemize}
\item Word2Vec-LSTM --- For this model, we generate Word2Vec embeddings of the opcodes, 
then train an LSTM model on the resulting sequence of embedding vectors.
\item BERT-LSTM --- This is the same as the Word2Vec-LSTM model, except that BERT is used to generate 
the embedding vectors, instead of Word2Vec.
\item Random Forest --- For this model, we train a Random Forest model directly on the opcode sequences. 
We obtain the feature vectors by truncating each sequence to a length~$L$, 
using the same values of~$L$ as for our HMM-CNN model. 
\item SVM --- As with the previous model, this model is also trained on the feature vectors obtained directly from the 
opcode sequences, but using an SVM classifier, instead of a Random Forest. 
\item HMM-RF --- This is similar to our HMM-CNN model, except that we apply a Random Forest classifier to the length~$7L$ feature vectors.
\item HMM-SVM --- This model is the same as the HMM-RF model, except that we use an SVM classifier in place of a Random Forest.
\item CNN --- For this model, we generate~$224\times 224$ images directly using the first $L$ opcodes of the malware samples. 
These images are then used to train a CNN classifier, as discussed in Section~\ref{subsec: training_procedure}, above.
\end{itemize}

Table~\ref{tab: acc_table} shows the accuracy and weighted F1-score obtained after testing each
of the above techniques on the same subset of the Malicia dataset as we used for our HMM-CNN experiments. 
We observe that the HMM-CNN slightly outperforms the HMM-SVM and HMM-RF, with Word2Vec-LSTM, Random Forest, 
and SVM models also performing reasonably well. Only the BERT-LSTM embedding achieved significantly lower accuracy 
and~F1 score, which is perhaps at least partially due to insufficient training data for the more complex BERT embeddings. 

\begin{table}[!htb]
\centering
\caption{Classification results for different techniques}\label{tab: acc_table}
\adjustbox{scale=0.85}{
\begin{tabular}{c|ccc}\midrule\midrule
\multirow{2}{*}{Technique} &  & \multicolumn{2}{c}{Validation} \\ \cline{3-4} \\[-2.25ex]
                 &  & Accuracy & F1-score\\ \midrule
Word2Vec-LSTM &  & 0.9714 & 0.9658\\
BERT-LSTM &  & 0.9181 & 0.9037\\
Random Forest &  & 0.9702 & 0.9668 \\
HMM-RF &  & 0.9758 & 0.9732\\
SVM &  & 0.9589 & 0.9535 \\
HMM-SVM &  & 0.9757 & 0.9727 \\
CNN &  & 0.9725 & 0.9727 \\
\textbf{HMM-CNN}&  & \textbf{0.9781} & \textbf{0.9778} \\
\midrule\midrule
\end{tabular}
}
\end{table}

Utilizing HMMs to generate feature vectors from opcode sequences clearly results in improvement in accuracy. 
Furthermore, the HMM-based models outperformed others when it came to classifying malware samples from 
some families that had limited representation in the dataset. For example, the HMM-based models classified 
malware samples of SecurityShield with an average accuracy of~87.5\%, while the non-HMM models
had an average accuracy of just~41.67\%.

\begin{figure}[!htb]
\centering
\begin{tikzpicture}[scale=0.65, every node/.style={scale=1.0}]
\pgfkeys{/pgf/number format/.cd,1000 sep={}}
\begin{axis}[
        width  = 0.85*\textwidth,
        height = 7.5cm,
        ymin=0.0,ymax=50.0,
        ytick={0, 10, 20, 30, 40, 50},
        major x tick style = transparent,
        ybar=5*\pgflinewidth,
        bar width=12.0pt,
        ylabel = {Training (hours)},
        symbolic x coords={Word2Vec-LSTM, BERT-LSTM, Random Forest, HMM-RF, SVM, HMM-SVM, CNN, HMM-CNN},
        xticklabels={Word2Vec-LSTM, BERT-LSTM, Random Forest, HMM-RF, SVM, HMM-SVM, CNN, HMM-CNN},
	y tick label style={
    		/pgf/number format/.cd,
   		fixed,
   		fixed zerofill,
    		precision=0},
        xtick = data,
        x tick label style={
        		rotate=60,
		font=\small,
		anchor=north east,
		xshift=5pt
		},
        enlarge x limits=0.1,
]
\addplot [fill=blue,opacity=1.00]
coordinates {
(Word2Vec-LSTM, 1.38)
(BERT-LSTM, 3.32)
(Random Forest, 0.64)
(HMM-RF, 24.91)
(SVM, 2.34)
(HMM-SVM, 26.26)
(CNN, 25.08)
(HMM-CNN, 48.83)
}; \label{plot_one}
\end{axis}
\begin{axis}[
        width  = 0.85*\textwidth,
        height = 7.5cm,
        ymin=0.0, ymax=0.10,
        ytick={0.0, 0.02, 0.04, 0.06, 0.08, 0.10},
        major x tick style = transparent,
        ybar=5*\pgflinewidth,
        bar width=12.0pt,
        axis y line*=right,
        bar shift=14pt,
        ylabel = {Testing per sample (seconds)},
        symbolic x coords={Word2Vec-LSTM, BERT-LSTM, Random Forest, HMM-RF, SVM, HMM-SVM, CNN, HMM-CNN},
        xticklabels={},
	y tick label style={
    		/pgf/number format/.cd,
   		fixed,
   		fixed zerofill,
    		precision=2},
        enlarge x limits=0.1,
        legend cell align=left,
        legend style={
                at={(0.11,0.79)},
                anchor=south,
                column sep=1ex
        },
]
\addlegendimage{/pgfplots/refstyle=plot_one}\addlegendentry{Training}
\addplot [fill=red,opacity=1.00]
coordinates {
(Word2Vec-LSTM, 0.0150)
(BERT-LSTM, 0.0753)
(Random Forest, 0.0010)
(HMM-RF, 0.0008)
(SVM, 0.0013)
(HMM-SVM, 0.0016)
(CNN, 0.0076)
(HMM-CNN, 0.0076)
};
\addlegendentry{Testing}
\end{axis}
\end{tikzpicture}
\caption{Training and testing timings}\label{fig:timings}
\end{figure}
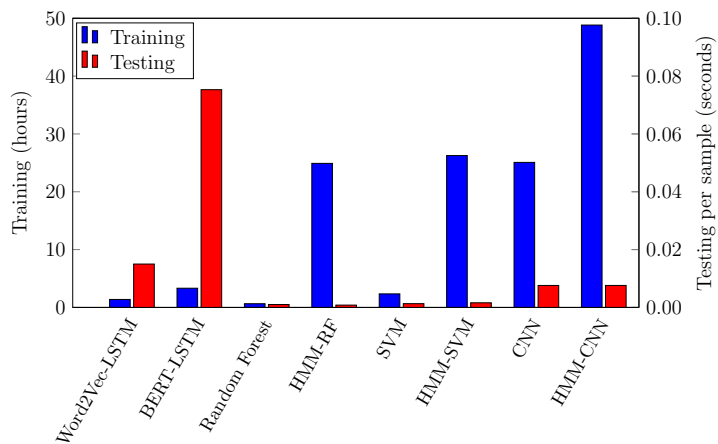

Table~\ref{tab:time_table} shows the average training and testing time of the tested techniques for the Malicia dataset. 
For emphasis, we give these same timing results in Figure~\ref{fig:timings}.

\begin{table}
    \centering
    \caption{Training and testing times for different techniques}\label{tab:time_table}
    \adjustbox{scale=0.85}{
    \begin{tabular}{c|cc}
        \midrule \midrule
        \multirow{2}{*}{Technique} & \multicolumn{1}{c}{\ Total training time\ } & \multicolumn{1}{c}{\ Testing time per sample\ } \\
        & (in hours) & (in seconds) \\
        \midrule
        Word2Vec-LSTM & \zz1.38 & 0.0150 \\
        BERT-LSTM & \zz3.32 & 0.0753 \\
        Random Forest & \zz\textbf{0.64} & 0.0010 \\
        HMM-RF & 24.91 & \textbf{0.0008} \\
        SVM & \zz2.34 & 0.0013 \\
        HMM-SVM & 26.26 & 0.0016 \\
        CNN & 25.08 & 0.0076 \\
        HMM-CNN & 48.83 & 0.0076\\
        \midrule
    \end{tabular}
    }
\end{table}

We observe that Word2Vec-LSTM and BERT-LSTM, 
which combine word embeddings with LSTM networks, require moderate training 
times 
and relatively high testing times. 
On the other hand, 
techniques incorporating HMMs required significantly longer training time, but maintain efficient testing 
times, while Random Forests are extremely efficient, with respect to both training and testing. 

\section{Conclusion and Future Work}\label{sect:conc}

In this paper, we analyzed a hybrid HMM-CNN model. Specifically, we derived features 
using HMMs, which were then converted into images, which were classified using advanced CNN architectures.
We found that our HMM-CNN model outperformed several comparable techniques on the same dataset. 
In contrast, techniques that did not use the HMM hidden state sequences as features performed measurably worse. 
This indicates that training an HMM and using it to uncover the hidden states can serve as a valuable feature engineering step. 
The hidden state sequence of HMMs are often used in Natural Language Processing (NLP) 
applications but, as far as the authors are aware, this approach
has only previously been applied to malware-related problems in our previous work~\cite{hmm-rf}. 
The results in this paper provide additional evidence that such NLP-inspired  approaches holds promise 
in the malware domain. Analogous approaches would be worth investigating in other domains as well.

There are many possible avenues for future work. For example, attempting to extend our
results for HMM-based models to various types of obfuscated malware---such as polymorphic
and metamorphic malware---would be an interesting challenge. 

The training times required for the HMM models was found to be large in comparison to other standard models.
Optimizing the HMM training times would be worthwhile future work.
For example, we could reduce the training times by reducing the length of the training opcode sequences.
We used all of the available training data to generate our HMMs, but the models would likely converge with far less
data. 

Utilizing hidden state sequences generated by HMMs in conjunction with Long Short-Term Memory (LSTM) 
networks is another possible area of future research work. Intuitively, leveraging LSTM to analyze hidden state sequences 
generated by HMMs could provide a more holistic view with respect to the temporal dynamics exhibited by malware. 
This approach holds the potential to enhance the accuracy and robustness of malware classification, as it leverages 
both the discriminative power of HMMs in identifying behavioral patterns and the feature learning capabilities of LSTMs.

Testing on larger and more challenging datasets could give us a more fine-grained view of the relative
strengths and weaknesses of hybrid models based on HMM-generated hidden state sequences. 
Ensemble modeling techniques are another area for potential future work. 
Generating multiple HMMs using random restarts could be used to create ensembles 
of HMM-RF and HMM-CNN models, potentially providing improved results.

\bibliographystyle{plain}
\bibliography{references.bib}

\end{document}